\documentclass[aps,prl,reprint]{revtex4-2}

\usepackage{graphicx} 
\usepackage{amsmath}
\usepackage{amssymb}  
\usepackage{bm} 
\usepackage{silence}
\usepackage[colorlinks=true,linkcolor=blue,urlcolor=blue,citecolor=blue]{hyperref}

\begin{abstract}
Increasing the L2 regularization of Deep Neural Networks (DNNs) causes a first-order phase transition into the under-parametrized phase -- the so-called onset-of learning. We explain this transition via the scalar (Ricci) curvature of the error landscape. We predict new transition points as the data complexity is increased and,   in accordance with the theory of phase transitions, the existence of hysteresis effects. We confirm both predictions numerically.  Our results provide a natural explanation of the recently discovered phenomenon of '\emph{grokking}' as  DNN models getting stuck in a local minimum of the error surface, corresponding to a lower accuracy phase. Our work paves the way for new probing methods of the intrinsic structure of DNNs in and beyond the L2 context. 
\end{abstract}

\begin{document}

\title{Phase Transitions between Accuracy Regimes in L2 regularized Deep Neural Networks}

\author{Ibrahim Talha Ersoy}
\email{talha.ersoy@uni-potsdam.de}
\affiliation{Complexity Science Group, Institute of Physics and Astronomy, University of Potsdam, Potsdam, Germany}

\author{Karoline Wiesner}
\email{karoline.wiesner@uni-potsdam.de}
\affiliation{Complexity Science Group, Institute of Physics and Astronomy, University of Potsdam, Potsdam, Germany}
\date{\today}

\maketitle

\textit{Introduction---}
L2 regularization is a common method used to avoid overfitting in different optimization tasks from simple linear regression  \cite{hoerl1970ridge} to regression and classification in artificial Deep Neural Networks (DNNs, i.e. Neural Networks with two or more hidden layers) \cite{Goodfellow-et-al-2016}. There is a rich phenomenology associated with L2 regularization, including phase transitions, non-trivial geometry of the loss landscape, and training dynamics that go well beyond the classical notion of overfitting control \cite{glassy,jacot2022}. For example, Ziyin and Ueda \cite{ziyin2023zeroth} show that NNs entering an over-regularized phase is equivalent to a second-order phase transition in the accuracy for one-hidden-layer Neural Networks and to a first-order phase transition for DNNs. 
Rubin, Seroussi and Ringel \cite{rubin2024grokkingorderphasetransition} have linked the phenomenon of \emph{grokking} to first order phase transitions using kernel methods. A general explanation for the observation of phase transitions in the training dynamics of DNNs, however, is lacking. 
In this Letter, we, first provide an  explanation of the onset-of-learning phase transition, as observed by Ziyin and Ueda \cite{ziyin2023zeroth}, in terms of the geometry of the NNs' error and loss surfaces. Second, we predict and confirm the existence of further phase transitions during regularization, beyond the onset-of-learning, i.e. when the data complexity is increased. And third, we provide an explanation for the phenomenon of \emph{grokking}, which is a substantial obstacle for fast convergence in the training of large DNNs \cite{demoss2025complexity,liu2023omnigrok}.
Our work unifies an information geometric and a regularization perspective, offering a framework for understanding the dynamics of DNN training. Specifically, our contributions include the reproduction and extension of Ziyin et al.'s  \cite{ziyin2023zeroth} results, focusing on DNNs, providing a geometric interpretation of the phase transitions they observed and the identification of transitions to under-parameterized phases corresponding to measurable geometric change-points in the error surface. We then observe additional transitions for more complex datasets, indicating the emergence of geometric substructures that represent learned features. We predict the existence of transitions beyond the onset-of-learning when the data complexity is increased. Again this transition comes with a geometric change-point in the error landscape. Focusing on the more common case of DNNs we confirm both hypotheses numerically measuring the scalar (Ricci) curvature as the regularization strength is varied.
Our findings provide a deeper understanding of phase transitions influenced by the interplay of regularization and data complexity. This work not only has potential practical implications for phenomena associated with over- and under-parametrization but also furthers our understanding of the geometry and structure of error landscapes and the notion of accuracy phases.  We also show that in L2 regularized settings hysteresis effects are responsible for delayed convergence. This happens when the model is initialized in a lower accuracy phase, or when the training trajectory passes through a lower accuracy phase. We reproduce \emph{grokking}-like behavior to link this to hysteresis and propose it will happen naturally when data-complexity is increased. 
\\
\textit{Setup---}The focus of our investigation is the L2-regularized loss function, defined as: 
\begin{align}\label{eq:lossfct}\mathcal{L}_{reg}(\mathbf{y},\mathbf{f}_{\boldsymbol{\theta}}(\mathbf{x})) = \mathcal{L}(\mathbf{y},\mathbf{f}_{\boldsymbol{\theta}}(\mathbf{x})) + \beta \|\theta\|^2
\end{align}, where \( \mathcal{L}(\mathbf{y},\mathbf{f}_{\boldsymbol{\theta}}(\mathbf{x})) \) is the (unregularized) error function (here the mean-squared error (MSE) ), and $\beta$ is the regularization strength. Both the error and the loss surface are manifolds in 
the Euclidean space spanned by the NN parameters  and the error and loss value, respectively.
In the following we study the curvature of these surfaces using the first $\mathrm{I}$ and second $\mathrm{II}$ fundamental form, i.e. the induced metric and the extrinsic curvature, of a manifold (for derivations, see Supplementary material). The so-called Gauss equation, i.e. a generalization of the Theorema Eggregium to higher dimensions, is used to calculate the Riemann tensor as follows:
\begin{align}
     R_{lijk} =  \mathrm{II}_{ik}\mathrm{II}_{jl}-\mathrm{II}_{ij}\mathrm{II}_{kl}~.
\end{align}
Using the inverse of the first fundamental form, we obtain the scalar (Ricci) curvature (for an alternative derivation see \cite{stephenson2021geometry}):  
\begin{align}
   R  &= \frac{1}{\| \nabla F \|} (tr(H)^2 - tr(H^2))  \nonumber \\ 
   &+ \frac{2}{\| \nabla F \|^2} \nabla l^T (H^2 - tr(H) H) \nabla l~.
\end{align}
Note that we obtain a formula that only requires gradients of the error, $\nabla l$, the Hessian, $H$, of the error surface and $ F(l, \theta_1, \ldots, \theta_d) =l- l(\boldsymbol{\theta}) = 0$.
Despite requiring a longer calculation the final result is  numerically more stable than the Gauss-Kronecker curvature. (The derivation of both curvature measures, as well as the relation to the Fisher Information is given in the supplementary material)

\begin{figure*}[t]
    \includegraphics[width=\textwidth]{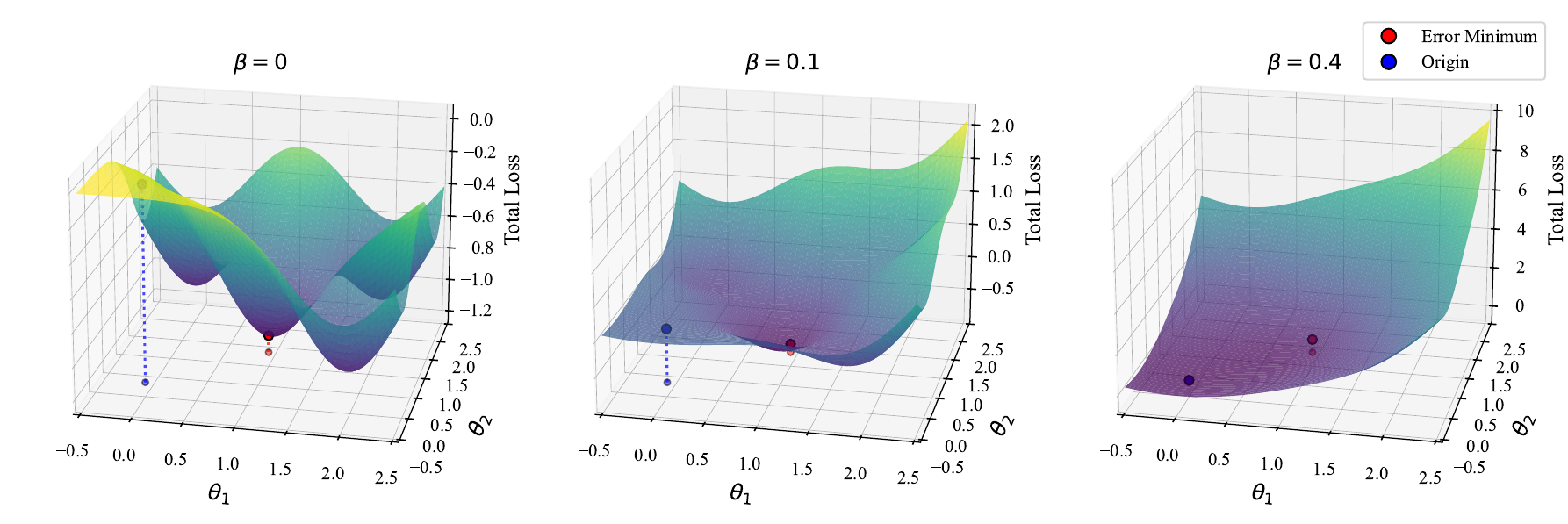}
    \caption{Sketch of a loss landscape with increasing L2 regularizer strength $\beta$. 
    The minimum of the error term (red) and the minimum of the L2 term at the origin (blue) are marked. 
    Increasing $\beta$ progressively smooths the landscape, eliminating local minima while shifting 
    the global minimum toward the origin.}
    \label{fig:loss_landscape}
\end{figure*}

\textit{Geometry of Learning---}Note, that increasing $\beta$  shifts the loss landscape's global minimum toward the origin (where all NN parameters $\theta$ equal zero) and away from the  minimum of the (unregularized) error landscape. The L2 regularizer is equivalent to a force field, shifting the model away from the minimum-error model. As $\beta$ becomes sufficiently large, the trained model suddenly transitions from a useful model to a ``trivial model'' with all parameters $\theta$ at or near zero. This transition, termed the \emph{onset-of-learning}, manifests as a first-order phase transition in DNNs, with the L2 norm of the parameters as the order parameter \cite{ziyin2023zeroth}. Fig.~\ref{fig:loss_landscape} illustrates the change of the loss function as the NN passes through an onset of learning, showing how increasing $\beta$ moves the error-minimizing model out of the region of parameters corresponding meaningful models.
We hypothesize, and subsequently confirm, that the model, with increasing regularization, traverses through a set of basins in the error landscape as it approaches the origin. Specifically, we hypothesize that:
(i) As a model is pushed out of an error basin in an L2 regularizer there is a phase transition in the model accuracy
(ii) the phase transition in the accuracy is caused by the model taking a jump on the error surface when $\beta$ reaches the onset-of-learning. If correct, this jump should be visible as a sudden change of curvature and of distance to the origin at the model location, as a function of $\beta$. 
(iii) the L2 regularizer can be used to pass the model from the highest accuracy parametrization to a trivial one , allowing us to control 'unlearning' to identify different accuracy regions and analyze the landscape with metrics like the Ricci scalar as the changes is accuracy happen. 
The number of transitions thus increases as the data-complexity is increased adding further basins to the error landscape. \\

\begin{figure*}[t]
    \includegraphics[width=\textwidth]{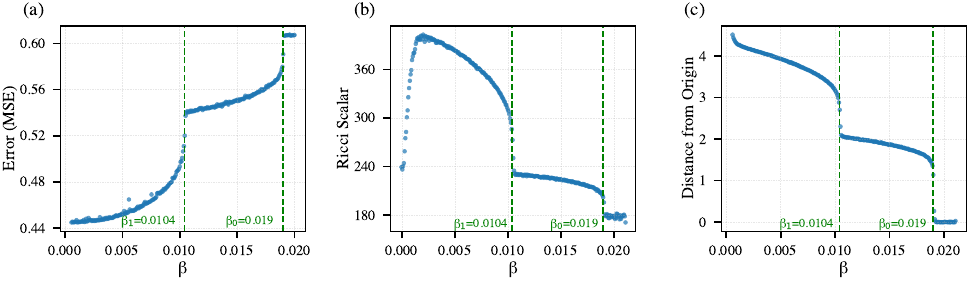}
    \caption{Phase transitions in a two hidden layer neural network trained on 2D Gaussian output.
    (a) Mean squared error, (b) Ricci scalar curvature, and (c) parameter space distance 
    all show transitions at $\beta_0$ (onset of learning) and $\beta_1$ (second transition point).}
    \label{fig:combined_results}
\end{figure*}

\textit{Results---}In Fig.~\ref{fig:combined_results} we present the experimental results for a two-hidden-layer NN, trained on 3D input  $\mathbf{X}$ and 2D output data $\mathbf{Y}$ sampled from a 5D Gaussian distribution $ (\mathbf{X},\mathbf{Y}) \sim \mathcal{N}(0, \Sigma_{xy})$ with $\Sigma_{xy}$ the joint covariance (see supplementary material for experimental details).

In Fig.~\ref{fig:combined_results}(a) we clearly see the onset-of-learning $\beta_0$ and an additional transition point $\beta_1$ in the accuracy / error curve. The transition at the onset-of-learning is a reproduction of a known result $\cite{ziyin2023zeroth}$ while the additional transition at $\beta_1$ supports hypothesis (iii). We can also clearly observe the curvature transition points in the model position as well as the sudden  jumps of the model on the error surface, measured in terms of distance to the origin, see Figs.~\ref{fig:combined_results}(b) and (c), as proposed in hypothesis (ii). A decrease in curvature as the model error increases. i.e. as the model 'unlearns', is also observed in Fig.~\ref{fig:combined_results}(b), which further supports hypothesis (i) as we would expect a flatting of the error surface when  moving out of an accuracy basin toward the origin (all parameters equal zero).

\begin{figure}[t]
\centering
  \includegraphics[width=\columnwidth]{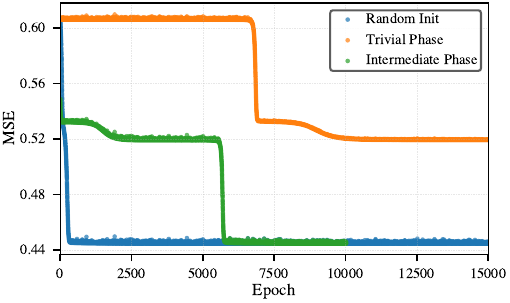}
  \caption{Mean-Squared-Error over Epochs for three models trained with different initialization  and small regularization strengths $\beta = 35e-5 < \beta_1$. While random initialization (blue) displays very fast convergence, initialization in the intermediate accuracy phase (green) and trivial accuracy phase (orange) lead to strongly delayed convergence due to hysteresis}
  \label{fig:hysteresis}
\end{figure}

\textit{Grokking---}
The phenomenon of \emph{grokking} is observable in  learning tasks of high complexity \cite{liu2023omnigrok} and in L2 regularized settings. \emph{grokking} refers to the phenomenon of an initial decrease in error, after which the model seems to have converged and displays no changes in accuracy anymore for a large number of epochs, until a second convergence sets in. , see Fig.~\ref{fig:hysteresis}, we reproduce precisely this behavior in our very small set-up. In this small set-up, \emph{grokking} is unlikely to occur. But we are able to artificially cause it, by initializing the model in the under-parametrized phase of the error surface and then train it with a very small regularizer ($\beta \ll \beta_1$). In this setting, the model is stuck for thousands of epochs in a low-accuracy phase until, finally, exiting it and reaching a high-accuracy phase ((orange curve in Fig.~\ref{fig:hysteresis})). We compare this to random initialization where the model quickly finds a high-accuracy phase on the error surface (blue curve in Fig.~\ref{fig:hysteresis}). The intermediate case, initializing the model in the intermediate accuracy phase, also exhibits \emph{grokking}, but the model escapes after fewer number of epochs (green curve in Fig.~\ref{fig:hysteresis}). 
In each case, we identify the model parameters associated with the respective phase using the experiments describe above (Fig.~\ref{fig:combined_results}). In other words, \emph{grokking} is nothing but a hysteresis effect associated with a first-order phase transition: the model is stuck in a local minimum, trapped due to the geometry of the error surface. 
We propose that, given high enough data-complexity, the likelihood of the model getting stuck in one of the local minima increases even with random initialization, leading naturally to this delayed convergence that manifests itself as \emph{grokking}. 

\textit{Conclusion---}
This study has demonstrated the existence of phase transitions in L2-regularized NNs beyond the onset of learning, particularly focusing on the existence and exploration of accuracy phases utilizing L2 regularization. Our investigation highlighted the changes of the error landscape's geometric structure as a model is pushed from high to lower accuracy. We have investigated the local geometry along a specific path that starts in the highest accuracy section we could find and ends at a parameter space section that corresponds to a trivial model. We can not deduce details on the global structure of the error landscape. Even though we have identified distinct accuracy phases, we can not yet tell whether all phases and change-points are detected.
However, we have established a framework connecting the transition phenomenon to accuracy phases and geometric metrics that will allow us to attack this open question. So far, having only demonstrating their existence, we can now follow up by further investigate the notion of error basins linked to the accuracy phases as well as better understand the global error geometry. We have also used our knowledge of first order phase transitions and hysteresis to reproduce \emph{grokking} in a minimal model to provide a simple mechanistic explanation for the phenomenon without usage of higher level terms like memorization, compression, etc. Our work leads us to hypothesize that increasingly complex learning tasks will have increasingly high probability of  displaying \emph{grokking}. 
 We leave a rigorous treatment including a precise bound on the numbers of local minima as well as schemes to avoid this hysteresis effect to future work.

\end{document}


\section{Geometry of the Model Space}

\subsection{Error and Loss Landscapes as Submanifolds in Model  Space }
Here, we derive and investigate the geometric properties of error and loss landscapes as $d$ dimensional submanifolds of a $d+1$ Euclidean ambient space, defined by a smooth error or loss function as the embedding.

The \emph{parameter space} $\Theta = \mathbb{R}^d$ is the high-dimensional space of all possible parameter configurations (weights and biases) for a given NN architecture. Each point $\boldsymbol{\theta} \in \Theta$ represents a specific set of weights and biases, i.e. a \emph{model}. The parameter space is equipped with the Euclidean metric, with
the previously defined L2 norm as the induced norm. We extend this parameter space to include an error dimension, resulting in $\Theta \times \mathbb{R} = \mathbb{R}^{d+1}$, which we equip with a global coordinate map $(l, \theta_1, \ldots, \theta_d)$, with $l \in \mathbb{R}$ denoting the error coordinate and the Euclidean metric $e_{ab} = \delta_{ab}$ in index notation with $a,b = 0,...,d$. We call this the \emph{model space}.
Given a smooth error function $\mathcal{L}(\mathbf{y},\mathbf{f}_{\boldsymbol{\theta}}(\mathbf{x}))$ evaluated on a dataset $D$, we can now define the error surface $\tilde{\mathcal{L}}$ as a Euclidean submanifold in the model space $\Theta \times \mathbb{R}$. The optimization (learning) process can be viewed as a trajectory on $\tilde{\mathcal{L}}$, where in practice some variation of gradient descent is used to locate the global minimum.
In general, the surface is highly non-convex and contain  saddle points and degenerate global minima  due to over-parametrization and the inherent symmetries of the NNs. Despite the complexity of the geometry, we can assert some minimal assumptions, such as the existence of basins that correspond to higher accuracy in the model output and that are formed by the features present in the dataset. 
We now introduce some concepts describing basic differential geometry of the error landscape.

\subsubsection{Fundamental Forms and Curvature of the Error Landscape}

Denoting the error function as $\mathcal{L}(\mathbf{y},\mathbf{f}_{\boldsymbol{\theta}}(\mathbf{x})) \equiv l(\boldsymbol{\theta}) $, we express the error surface in parametric form as:
\begin{equation}
f(\theta_1, \ldots, \theta_d) = (l(\boldsymbol{\theta}), \theta_1, \ldots, \theta_d).
\label{eq:parametric}    
\end{equation} 
Alternatively, we can define it using an implicit function formulation:
\begin{equation}
    F(l, \theta_1, \ldots, \theta_d) =l- l(\boldsymbol{\theta}) = 0
\label{eq:implicit}    
\end{equation}

\begin{figure}[t]
    \centering
    \includegraphics[width=\columnwidth]{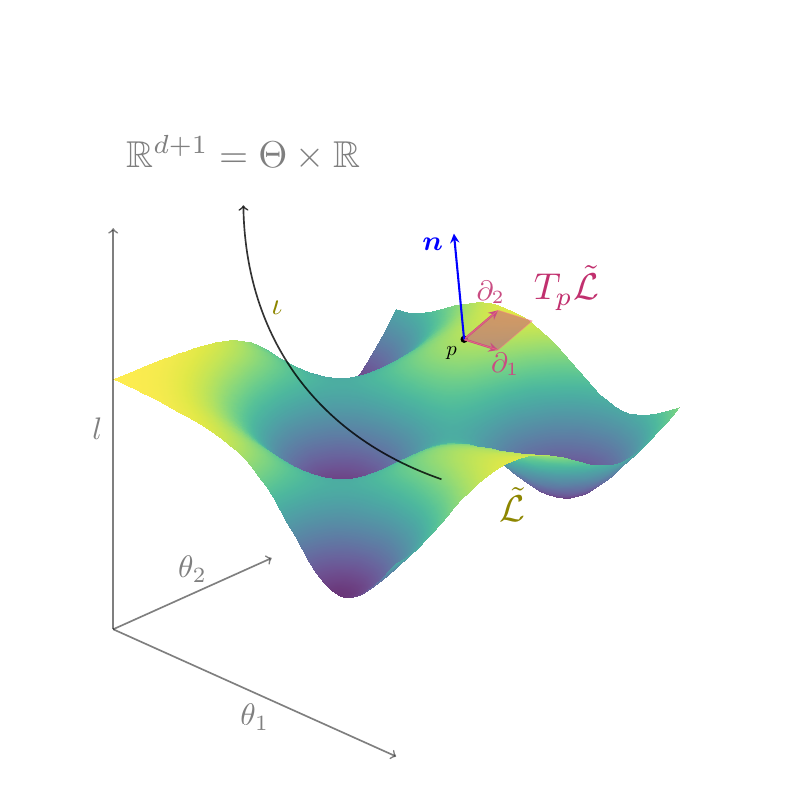}
    \caption{Sketch of an error surface $\tilde{\mathcal{L}}$ as a $d$-dimensional submanifold of the Euclidean ambient space $\mathbb{R}^{d+1}$, consisting of the parameter space $\Theta$ and the error values. The tangent space $T_p \tilde{\mathcal{L}}$ at $p \in \tilde{\mathcal{L}}$ is spanned by $\partial_i = \frac{\partial}{\partial\theta_i}$.}
    \label{fig:submnf}
\end{figure}

We have the inclusion map for the submanifold in the ambient space  \ref{fig:submnf}:
\begin{equation}
\iota: \tilde{\mathcal{L}} \to \Theta \times \mathbb{R}.
\end{equation}
Let $T_p L$ be tangent space at some point \(p \in \tilde{\mathcal{L}}\). For elements \(v, w \in T_p L\), the scalar product \(\langle v, w \rangle_p\) on \(T_p  \tilde{\mathcal{L}}\) is given by:
\begin{equation}
\langle v, w \rangle_p = (d \iota)_p(v)(d \iota)_p(w).
\end{equation}

This is the \emph{first fundamental form}  in  coordinate-free form, also known as the \emph{pull-back} of the Euclidean metric in $d+1$-dimensional space to the submanifold. 
Given the coordinates $ (l,\theta_1,...,\theta_d)$ we get the basis of the tangent space $ \left( \frac{\partial}{\partial l}, \frac{\partial}{\partial\theta_1},...,\frac{\partial}{\partial\theta_d} \right) $.
With the Euclidean metric  $e_{ij} = \delta_{ij}$ with $i,j = 1,...,d $,
we get the components $g_{ij}$ of the induced metric:
\begin{equation}
g_{ij} = \delta_{ij} + \frac{\partial l}{\partial \theta_i} \frac{\partial l}{\partial \theta_j}~.
\label{eq:first_fund}
\end{equation}
The first fundamental form gives us the scalar product and, hence, can be used to calculate lengths, angles and volumes on the submanifold.
We now also derive the second fundamental form because it can be used to obtain the intrinsic curvature of the submanifold. 

To derive the second fundamental form, that measures the extrinsic curvature, we need to consider the normal space, which is one-dimensional, as we have a d-dimensional submanifold embedded in a (d + 1)-dimensional ambient space. 
With the implicit formulation given in Eq.~\ref{eq:implicit},  denoting $\partial_i = \frac{\partial}{\partial\theta_i}$ and  $\| \nabla F \|\ = \sqrt{1+\sum_{i=1}^d |\partial_il|^2}$, the normal vector field $\mathbf{n}$ is given by:
\begin{align}
        \mathbf{n} &= \frac{-\nabla F}{\| \nabla F \|} = \frac{1}{\| \nabla F \|} \left(\partial_0 F, \partial_1F, \cdots , \partial_d F \right) \nonumber \\ 
        &= \frac{1}{\| \nabla F \|} \left(1,- \partial_1l, \cdots , -\partial_d l \right)
\end{align}

With the parametrization $f$ given in Eq.~\ref{eq:parametric} and with $f_{ij} = \partial_i \partial_j f = (\frac{\partial^2 l }{\partial \theta_i \partial \theta_j}, 0, \cdots , 0) $ the second fundamental form $\mathrm{II}_{ij}$ is:
\begin{align}
    \mathrm{II}_{ij} & =\mathbf{n} \cdot  f_{ij} = \frac{1}{\| \nabla F \|}  \left(1, -\partial_1l, \cdots , -\partial_d l \right )  \nonumber \\
    & \times \left(\frac{\partial^2 l }{\partial \theta_i \partial \theta_j}, 0, \cdots , 0 \right)  = \frac{H_{ij}}{\| \nabla F \|}~,
\end{align}
where $ H_{ij} $ is the Hessian of the loss function, scaled by the norm of the gradients plus one.
This is a useful relation as the gradients are a key component in training and thus easily available for each trained NN. The Hessian is more expensive to obtain but cheap enough for the NNs we are investigating. 
As a symmetric bilinear form, the second fundamental form can be diagonalized at any point in the submanifold with  eigenvectors and -values $v^{(i)}, \kappa^{(i)}$ .  The eigenvalues are referred to as principal curvatures.
We can easily see that the eigenvectors of the second fundamental form and the Hessian are the same:
\begin{equation}
   \kappa^{(i)}  v^{(i)} = \mathrm{II}  \cdot  v^{(i)}  =  \frac{-1}{\| \nabla F \|} H \cdot v^{(i)}  = \frac{-1}{\| \nabla F \|}  \tilde{\kappa}^{(i)} v^{(i)}  
\end{equation}
We get the relation between the principal curvatures $\kappa^{(i)}$ and the Hessian eigenvalues  $\tilde{\kappa}^{(i)}$:
\begin{equation}
    \kappa^{(i)}  = \frac{-1}{\| \nabla F \|}  \tilde{\kappa}^{(i)}  
\end{equation}
Consequently, the principal curvatures can be readily obtained from the eigenvalues of the Hessian and the gradients of the error function, $\partial_i l$. With the determinant of the second fundamental form and the gradients, the Gauss-Kronecker curvature can be obtained:
\begin{equation}
    K = \frac{\det(\mathrm{II})}{\det(g)}= \frac{1}{\| \nabla F \| ^d}\frac{\prod_{i=1}^d \tilde{\kappa}^{(i)}}{\det(g)}~.
\label{eq:GK-curv.}    
\end{equation}
The Problem with the Gauss-Kronecker curvature is the product over the large number of eigenvalues. Most eigenvalues are close to zero. The curvature can thus only be obtained by introducing a cutoff. Even then we get numerically unstable results. We thus derive a more stable curvature measure that scales better in higher dimensions.
I.e. we use the second fundamental form to derive the Riemann-curvature.  
For Euclidean submanifolds the Gauss Equation relating the second fundamental form (extrinsic curvature) to the Riemann tensor (intrinsic curvature) gives us the following expression for the Riemann-curvature. We start with a generalization of the Theorema Eggregium for arbitrary dimensions, also called Gauss-Equation. It related the Riemann Curvature to the second fundamental form. For Euclidean submanifolds it is given as:

\begin{equation}
    R_{lijk} =  \mathrm{II}_{ik}\mathrm{II}_{jl}-\mathrm{II}_{ij}\mathrm{II}_{kl}
\end{equation}
The Riemann tensor can be contracted to get the Ricci tensor. In the Einstein summation notation Ricci curvature tensor  is then given with the inverse of the metric $g^{ij}$:
\begin{equation}
    R_{ij} = R^{k}_{ikj} = g^{kl}\Pi_{kl} \Pi_{ij} - \Pi_{ik} g^{km} \Pi_{mj}
\end{equation}
Contracting the indices again we get the Ricci scalar:
\begin{align}
   R &= g^{ij} R_{ij}  = g^{kl}\Pi_{kl} g^{ij} \Pi_{ij} -  g^{ij}  \Pi_{ik} g^{km} \Pi_{mj}  \nonumber \\ 
   &= \tilde{H}^2 - |\Pi|^2,
\end{align}
with the mean curvature  $\tilde{H} = g^{ij}\Pi_{ij}$ and the squared norm of the second fundamental form $|\Pi|^2 = g^{ij} g^{kl} \Pi_{ik}\Pi_{jl}$.
We use the the Sherman-Morrison Formula to give inverse of the metric:
\begin{equation}
    g^{ij} = \delta^{ij} - \frac{\partial_i l \partial_j l }{\| \nabla F \|^2}
\label{eq:inverse}    
\end{equation}
With  $\Pi_{ij} = \frac{H_{ij}}{\| \nabla F \|} $ Plugging this into the mean curvature we get: 
\begin{align}
    \tilde{H}^2 &= \frac{1}{\| \nabla F \|^2}   \left[ \left(\delta^{ij} - \frac{\partial_i l \partial_j l}{\| \nabla F \|^2}\right) H_{ij} \right]^2   \nonumber \\
    &= \frac{1}{\| \nabla F \|^2}\left[ tr(H) - \frac{\nabla l ^T H \nabla l}{\| \nabla F \|^2} \right]^2  \nonumber \\
   &=
   \frac{1}{\| \nabla F \|^2} \left( tr(H) \right)^2 -2\frac{ tr(H) \nabla l ^T H \nabla l}{\| \nabla F \|^4} \nonumber \\
    &+ \frac{\left( \nabla l ^T H \nabla l \right)^2}{\| \nabla F \|^6} 
\label{eq:mean_curv}    
\end{align}  
For the second term we get:
\begin{align}
  \frac{1}{\| \nabla F \|^2} g^{ij} g^{kl} \Pi_{ik}\Pi_{jl} &= \frac{1}{\| \nabla F \|^2} (\delta^{jk}\delta^{il} \nonumber \\
 -\delta^{jk} \frac{\partial_i  l \partial_l l}{\| \nabla F \|^2} -\delta^{ij} &\frac{\partial_j  l \partial_k l}{\| \nabla F \|^2}  +\frac{\partial_j  l \partial_k l \partial_i l \partial_l l}{\| \nabla F \|^4})H_{jk} H_{il} \nonumber \\
   = \frac{tr(H)^2}{\| \nabla F \|^2}   - &\frac{2  \nabla l ^T H^2 \nabla l }{\| \nabla F \|^4} + \frac{\left( \nabla l ^T H \nabla l \right) ^2}{\| \nabla F \|^6} \nonumber \\
\label{eq:squared_second}    
\end{align}

With Eq.s~\ref{eq:mean_curv} and \ref{eq:squared_second} we get the final expression:
\begin{align}
   R  &= \frac{1}{\| \nabla F \|} (tr(H)^2 - tr(H^2))  \nonumber \\ 
   &+ \frac{2}{\| \nabla F \|^2} \nabla l^T (H^2 - tr(H) H) \nabla l
\end{align}
The Ricci curvature in the experimental results is obtained using this expression.

\subsection{Curvature and Fisher Information} 
\graphicspath{{images/}}

\subsubsection{Metric of the Error Submanifold and the Fisher Information}
\label{app:likelihood_fisher}

Consider a Deep Neural Network (DNN) with parameters $\mathbf{\theta}$, input $\mathbf{x}$, and output $\mathbf{f}(\mathbf{x}; \boldsymbol{\theta})$ and the target variable $\mathbf{y}$ $\mathbf{x}$:
We have a mean-squared error loss $\operatorname{MSE}(\mathbf{y},\mathbf{f}_{\boldsymbol{\theta}}(\mathbf{x}))= \frac{1}{N} \sum_{i=1}^N (\mathbf{y}_i - \mathbf{f}_{\boldsymbol{\theta}}\left(\mathbf{x}_i) \right)^2 $and Gaussian error with $\sigma^2$ is the variance. The likelihood of observing $\mathbf{y}$ given $\mathbf{x}$ and $\mathbf{\theta}$ is given by:

\begin{align}
p(\mathbf{y}|\mathbf{x};\boldsymbol{\theta}) &= 
\frac{1}{(2\pi\sigma^2)^{n/2}} \exp\left(-\frac{l}{2\sigma^2} \right) \nonumber \\
&= \frac{1}{(2\pi\sigma^2)^{n/2}} \exp\left(-\frac{1}{2\sigma^2} \|\mathbf{y} - \mathbf{f}(\mathbf{x}; \boldsymbol{\theta})\|^2\right),
\label{eq:likelihood}
\end{align}

where $n$ is the dimensionality of $\mathbf{y}$. The log-likelihood is:

\begin{align}
\log p(\mathbf{y}|\mathbf{x}; \boldsymbol{\theta}) &= -\frac{n}{2} \log(2\pi\sigma^2)  \nonumber \\
&- \frac{1}{2\sigma^2} \frac{1}{N} \sum_{i=1}^N (\mathbf{y}_i  - \mathbf{f}_{\boldsymbol{\theta}}\left(\mathbf{x}_i) \right)^2.
\end{align}

\subsubsection{Fisher Information Matrix}
The Fisher Information Matrix (FIM) quantifies the amount of information that the observable random variable $\mathbf{y}$ carries about the parameters $\boldsymbol{\theta}$. The FIM is defined as the expected outer product of the score function:

\[
\mathbf{I}(\boldsymbol{\theta}) = \mathbb{E}\left[\mathbf{S}(\boldsymbol{\theta}) \mathbf{S}(\boldsymbol{\theta})^\top\right],
\]

where the score function $\mathbf{S}(\boldsymbol{\theta})$ is the gradient of the log-likelihood with respect to $\boldsymbol{\theta}$:

\[
\mathbf{S}(\boldsymbol{\theta}) = \frac{\partial \log p(\mathbf{y}|\mathbf{x}; \boldsymbol{\theta})}{\partial \boldsymbol{\theta}}.
\]

For the likelihood (Eq.~\ref{eq:likelihood}), the score function is:

\begin{equation}
\mathbf{S}(\boldsymbol{\theta}) = \frac{1}{\sigma^2} (\mathbf{y} - \mathbf{f}(\mathbf{x};\boldsymbol{\theta})) \frac{\partial \mathbf{f}(\mathbf{x}; \boldsymbol{\theta})}{\partial \boldsymbol{\theta}}=\frac{1}{\sigma^2} \frac{\partial l}{\partial \boldsymbol{\theta}}
\end{equation}

Substituting the score function into the definition of the FIM, we obtain:

\begin{align}
    \mathbf{I}(\boldsymbol{\theta}) = \mathbb{E}\left[\frac{1}{\sigma^4} (\mathbf{y} - \mathbf{f}(\mathbf{x}; \boldsymbol{\theta}))^2 \frac{\partial \mathbf{f}(\mathbf{x}; \boldsymbol{\theta})}{\partial \boldsymbol{\theta}} \frac{\partial \mathbf{f}(\mathbf{x}; \boldsymbol{\theta})}{\partial \boldsymbol{\theta}}^\top\right].
\end{align}

With $l$ the loss over all of the dataset we can write this in components as:
\begin{equation}
    \mathbf{I}_{ij} = \frac{1}{\sigma^2}  \frac{\partial l}{\partial \theta_i} \frac{\partial l}{\partial \theta_j}
\end{equation}

This is precisely the second term of the Eq.~\ref{eq:first_fund}  under the assumptions of Gaussian noise and squared error function times the noise parameter. The FIM gives us the geometric properties of the log-likelihood surface, while our metric is the metric of the error submanifold. The relation holds for normal error and quadratic error terms. Both spaces are locally isometric as they give the same curvature values as can easily be checked.

\section{Varying  Data-Complexity and Variational Autoencoders}
\subsection{Only one transition on 1D Gaussian Output, i.e. the Lowest Complexity Data-Set}
We start this section with the simplest dataset we can think of, i.e. a model trained on a 3D Gaussian distribution with the input being the 2D x-marginal and the output the correlated 1D y-marginal.

\begin{figure*}[t]
    \includegraphics[width=\textwidth]{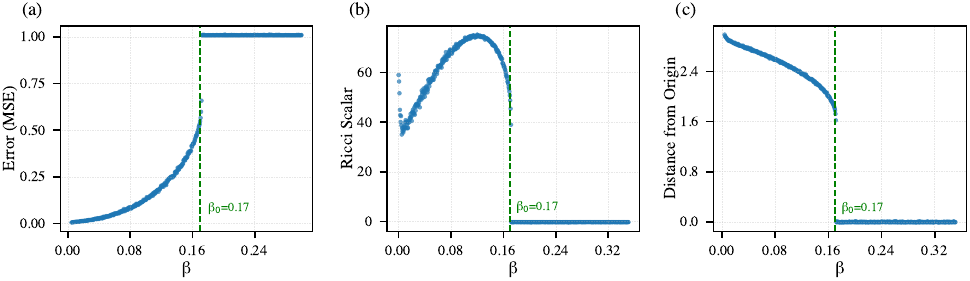}
    \caption{Phase transition in a two hidden layer neural network trained on 1D Gaussian output.
    (a) Mean squared error, (b) Ricci scalar curvature, and (c) parameter space distance 
    all show one transition at $\beta_0$ (onset of learning).}
    \label{fig:1D-combined_results}
\end{figure*}

The phenomenology here is very similar to the 2D case presented in the main section, with the difference that only one transition can be observed, i.e. the onset-of-learning. Again we see the jump in the error Fig.~\ref{fig:1D-combined_results}(a) and the drop in curvature Fig.~\ref{fig:1D-combined_results}(b). We also see the jump in the parameter space as Fig.~\ref{fig:1D-combined_results}(c). It is indicative of the relation between model data-complexity and number of transition points. 

\subsection{Further Transitions for Higher Complexity Task, i.e. Image Classification}

\begin{figure}[t]
    \centering
    \includegraphics[width=\columnwidth]{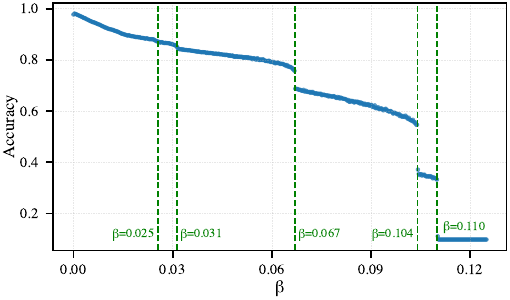}
    \caption{Accuracy of a NN with two hidden layers, trained on the MNIST dataset for image classification. The metrics are evaluated for increasing L2 regularization strength $\beta$. The transition points are again determined with some change-point detection algorithm and marked as $\beta_i$}
    \label{fig:MNIST}
\end{figure}

We also looked at a vastly higher dimensional case, i.e. a classifier. The setup is very similar, but instead of an MSE error we use a cross entropy error 
\(\operatorname{CE}(\mathbf{y},\mathbf{f}_{\boldsymbol{\theta}}(\mathbf{x}))\) and again add a regularizer with strength $\beta$ train a model with a total loss \( \mathcal{L}_{reg}(\mathbf{y},\mathbf{f}_{\boldsymbol{\theta}}(\mathbf{x}))= \operatorname{CE}(\mathbf{y},\mathbf{f}_{\boldsymbol{\theta}}(\mathbf{x}))+ \beta \|\theta\|^2\), successively  increasing the regularizer strength. 
The dataset is the MNIST dataset that consists of images of handwritten numerals from zero to ten and the corresponding labels. In Fig.~\ref{fig:MNIST} we can clearly see the onset of learning, again labeled $\beta_0$ and four additional transition points, in the accuracy (percentage of correct classifications) against $\beta$.
The input has 784 nodes giving us a problem that is not only fundamentally different as it is a classification task, but also quite far from being a mere toy model as the previous 1- and 2- dimensional Gaussian datasets. 

\subsection{Second order Phase Transitions with one Hidden Layer}

When we decrease the number of hidden layers to just one, the number of transitions remains the same but the transitions become continuous, in perfect alignment with what we would expect for second order phase transitions. 
\begin{figure*}[t]
    \includegraphics[width=\textwidth]{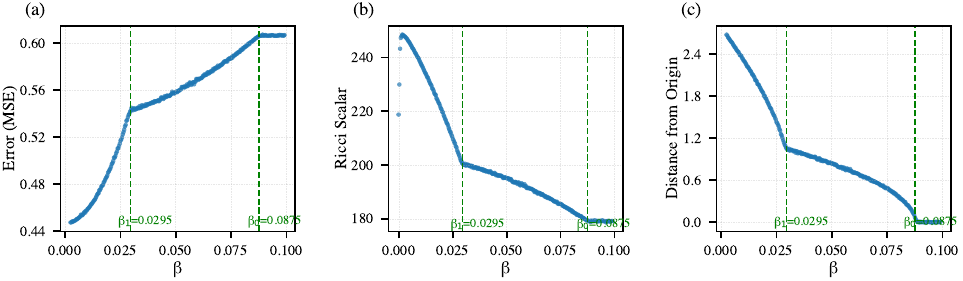}
    \caption{Phase transitions in a two hidden layer neural network trained on 2D Gaussian output.
    (a) Mean squared error, (b) Ricci scalar curvature, and (c) parameter space distance 
    all show transitions at $\beta_0$ (onset of learning) and $\beta_1$ (second transition point).}
    \label{fig:1hidden-combined_results}
\end{figure*}
Again we see two transition points and in general similar behavior to the 2-hidden-layer model trained on a 2D Gaussian output. The main difference is the lack of a jump in the parameter space. We leave further investigation for later work as this is not our main subject of interest.

\begin{figure}[t]
    \centering
    \includegraphics[width=\columnwidth]{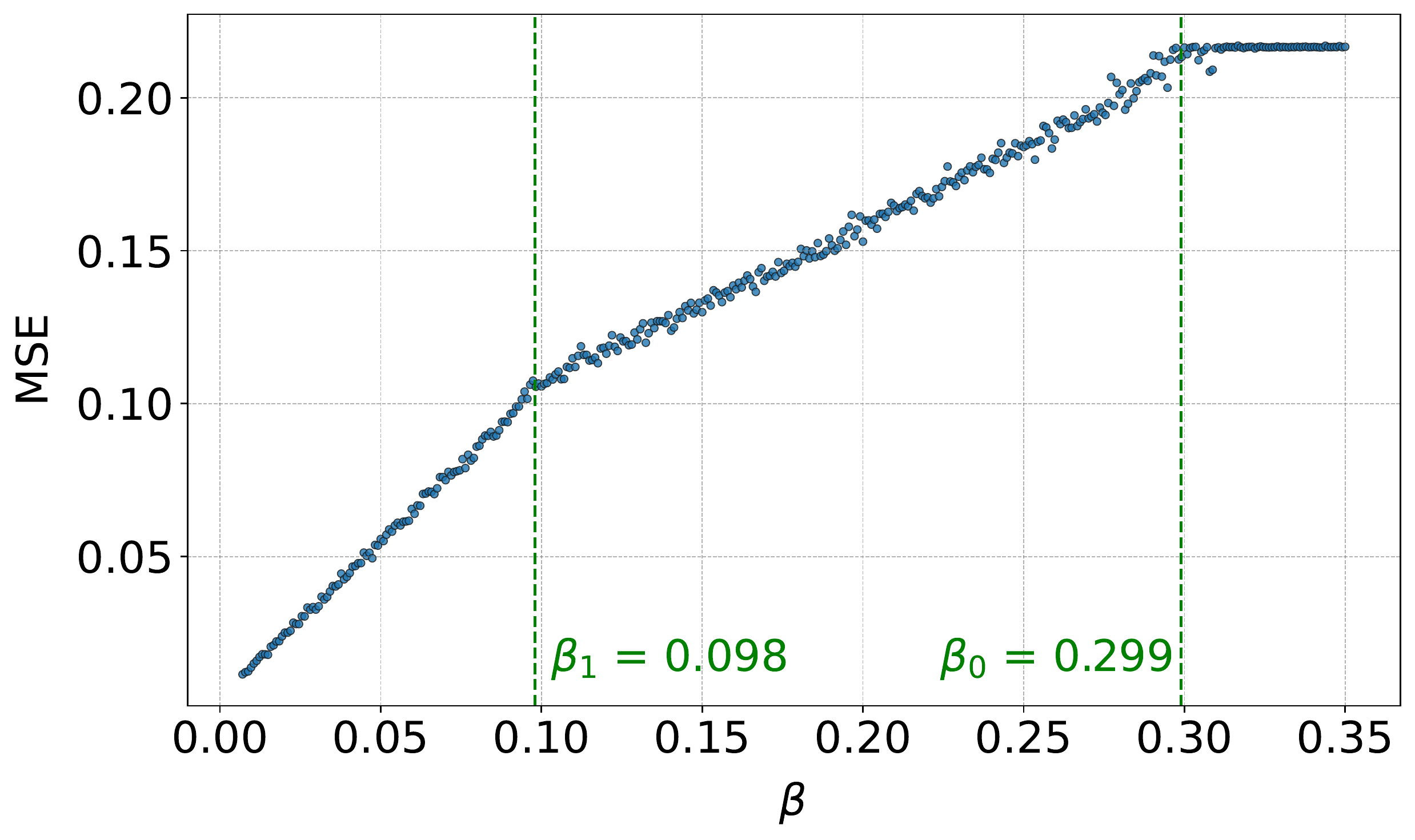}
    \caption{MSE for increasing regularizer strength $\beta$. The regularizer is the KL-divergence term (Eq.~\ref{eq:VAE}) of the VAE like model instead of the L2}
    \label{fig:VAE}
\end{figure}

We finally look at a different set-up without an L2 regularizer.
We trained a VAE like model with a loss of the form:
\begin{equation}
    \mathcal{L} = \operatorname{MSE}(\mathbf{y},\mathbf{f}_{\boldsymbol{\theta}}(\mathbf{x})) + \beta D_{KL}(p|q)
    \label{eq:VAE}    
\end{equation} 
with $p_{\theta}(l|x)$ the encoding map and $q\propto \mathcal{N}(0,\mathbf{1})$ white noise.
The model is trained on the same dataset $(\mathbf{X},\mathbf{Y}) \propto \mathcal{N}(0, \Sigma_{xy})$, with the same error function as before. Increasing the $\beta$ parameter pushes the model away from the error minimum toward the some trivial model that outputs noise. This gives us a mechanisms that is very different from the L2. It introduces a second term to the loss function that has a minimum at a model that has its minium defined by another model representation that is not defined by a location in the parameter space but by the neural network that has a latent representation that is closest to white noise. This gives us a more costly and less direct method of manipulating the model. However it gives a comparable manipulation of the model. We start at the MSE minimum determined by the data and gradually shift the effective minimum and thus the model toward a trivial model. 
The KL-divergence measures a sort of distance, but it is not the euclidean distance to the origin in parameter space but the distance to the trivial model.
In the error(MSE)-$\beta$ plot (Fig.~\ref{fig:VAE}) we see very similar phenomenology to the L2 regularized models. This is to be expected as we argued that the transition is just a consequence of the model traversing the geometric change-points in the error landscape. Given that we have the same error function and the same dataset we expect the error landscape to have a very similar (not exactly the same as the parametrization and the precise model is a bit different) geometry. 

\section{Experimental Setup}

\subsection{Setup 1: Simple Neural Network with L2 Regularization}
\label{app:experiment1}

In the first experimental setup, a simple feedforward neural network was implemented with a varying regularization parameter denoted by $\beta$. The main experiment consists in a loop where a network is trained tested and then trained again with an increasing $\beta$ value. A specific $\beta$ interval and the number of networks to be trained is set in the beginning. There are two possible modes. The first is the annealing mode where the parameters of the previous model are taken as the initialization of the following model. Without annealing a new network is trained for each value. 

\subsubsection{Architecture}
The network consists of an input layer, $n$ hidden layers, and an output layer. The input layer has two or more dimensions corresponding to the features, and the output layer has one or more dimensions representing the target variable. For the main experiments the hidden layer contains 15 neurons and utilizes the Sigmoid activation function.
The choice of width is just taken by running a number of trials. Above 15 there is no significant improvement of the performance. Decreasing it for the non-annealing mode has the effect that there is an overlap of phase. Close to transition point some of the models with lower than critical $\beta$ converge to the highest possible accuracy while others get stuck at lower accuracy regimes.

\subsubsection{Data Generation}
We used Gaussian data with zero mean for all of the experiments. The covariance matrices are generated by taking a positive definite diagonal matrix of the right dimensionality and rotate it until the x- and y-distributions are highly correlated. 
For generalizability of the setup to  higher dimension we use the Cholesky decomposition for sampling. The sampling process begins with generating a set of raw data samples from a standard multivariate normal distribution, which includes generating random values with zero mean and unit variance. To shape these uncorrelated samples into a distribution with the desired properties, we apply the Cholesky decomposition of the specified covariance matrix. This transformation adjusts the samples so they reflect the variability and correlation defined by the covariance. Finally, we add the mean vector (zero in our experiments) to each sample, ensuring that the samples are centered around the specified mean values. This method ensures the resulting samples faithfully represent the intended multivariate Gaussian distribution characterized by the given mean and covariance. The first two dimensions are taken as input $(x_1,x_2)$ and the  last dimension $y$ as the output. The sample size for training and testing is $N=10^4$ each. This is chosen by trying a number of different orders of magnitude. Above this the performance does not improve significantly. 
To test the robustness of the findings we used the following scheme to generate other higher dimensional covariance matrices. We created a diagonal matrix of the desired shape and rotated in the given axis by random very small increments until we get some covariance matrix with a high enough correlation strength (close to one) between the $x$ and $y$ dimensions. This approach makes the scheme of choosing a distribution relatively random and ensures that the resulting distribution is positive definite. For the 2D and 3D experiments we picked the first of the randomly generated covariances.

\subsubsection{Training Process}
The model was trained for a specified number of epochs, with a varying $\beta$ value in each iteration to examine the impact of regularization on performance. The training utilized an optimizer (configured as either SGD, Adam, or AdamW) to minimize the loss function, which combined the mean-squared error (MSE) and the L2 regularization term.
The AdamW algorithm gives the best performance with faster convergence. The resulting epoch outputs are identical to the SGD ones. We thus argue that the choice of algorithm is mostly irrelevant to the results. We only look at the trained model and analyze the properties of the error landscape section and not the scheme to get to this point. 
\\
During the training, the model's performance was evaluated, and metrics such as MSE were recorded. 
After the last epochs the model parameters, the gradients and the eigenvalues of the Hessian matrix were evaluated and saved. Also the full hessian matrices were calculated for each trained model and saved in an array to be evaluated in curvature calculations.

\subsubsection{Hysteresis and Grokking}
The experimental setup for the hysteresis experiments is basically the same as the main experiment. But instead of multiple runs we just pick one sub-critical beta value, i.e. one that lies far below the transition points associated with the lower accuracy regimes. We then train three different models. The first one is randomly initialized and then trained for six-thousand epochs. For the second run we check the accuracy regimes from the main experiment and pick the parametrization of a model in the trivial phase to initialize a second model and repeat the training. Lastly we pick the model from the intermediate accuracy phases that is closest to the transition point and take its parameters to initialize a model for the last run and train it, again for six-thousand epoch and again with the same choice of hyper-parameters.

\subsubsection{Evaluation}
The evaluation was done in a very straight forward way. There is no transformation or processing done to the data and its just plotted as viewed in chapters (3) and (4), except for the Gauss-Kronecker calculation. We introduce a cut-off at $10^{-10}$ for the hessian eigenvalues and throw out all values below to avoid a zero determinant. Besides this we calculate the log-determinant and get the exponential afterwards for some numerical stability.

\subsection{Setup 2: Variational Autoencoder (VAE)}
\label{supp.:VAE}

The second experimental setup involved a Variational Autoencoder (VAE)-type setup. The main difference here is that, unlike the standard VAE our model is not trained to regenerate the input $\mathbf{x}$, but to produce the corresponding output $\mathbf{y}$. The latent representation is used to regularize the model and probe the range from no regularization to over-regularization, i.e. the point where the regularization is so strong it shifts the model into a trivial regime as in the L2 setup. The regularizer term in Eq.~\ref{eq:VAE}  takes the latent representation as the mean and variance of a Gaussian distribution and calculates Kullback-Leibler divergence with white noise. For high enough $\beta$ the minimization  of the regularizer term becomes the main objective leading to the latent representation becoming trivial. 

\subsubsection{Architecture}
This VAE consists of an encoder and a decoder. The encoder takes the input data and transforms maps into a latent space with 2 dimensions. It outputs both the mean and log variance of the latent variables, allowing for sampling during training through the reparameterization trick. It interprets the latent space as a parametrization of a Gaussian map and calculates the KL-divergence with white noise. For high enough $\beta$ the effective minimum shifts to the minimum of the KL-divergence term, forcing the model out of the error minimum and outputting a trivial model. 

\subsubsection{Data Generation}
To train and test the VAE, synthetic Gaussian data was generated same as in the first setup.

\subsubsection{Training Process}
Each training session involved adjusting the hyperparameter $\beta$. The basic set-up is analogous to the L2 experiments. One can choose an annealing approach or a non-annealing one. The outputs here only consist of the MSE and the KL-divergence values and the model parameters for each $\beta$ on the test data after each final epoch, as well as all epoch outputs.  
As in the L2 set-up we used sigmoid activations.